# Conditional Semi-Supervised Data Augmentation for Spam Message Detection with Low Resource Data

Ulin Nuha*, Chih-Hsueh Lin

Department of Electronic Engineering, National Kaohsiung University of Science and Technology, Kaohsiung, 80778, Taiwan

Email: i110152118@nkust.edu.tw; cslin@nkust.edu.tw

Corresponding author*

**Abstract**

Several machine learning schemes have attempted to perform the detection of spam messages. However, those schemes mostly require a huge amount of labeled data. The existing techniques addressing the lack of data availability have issues with effectiveness and robustness. Therefore, this paper proposes a conditional semi-supervised data augmentation (CSSDA) for a spam detection model lacking the availability of data. The main architecture of CSSDA comprises feature extraction and enhanced generative network. Here, we exploit unlabeled data for data augmentation to extend training data. The enhanced generative in our proposed scheme produces latent variables as fake samples from unlabeled data through a conditional scheme. Latent variables can come from labeled and unlabeled data as the input for the final classifier in our spam detection model. The experimental results indicate that our proposed CSSDA achieves excellent results compared to several related methods both exploiting unlabeled data and not. In the experiment stage with various amounts of unlabeled data, CSSDA is the only robust model that obtains a balanced accuracy of about 85% when the availability of labeled data is large. We also conduct several ablation studies to investigate our proposed scheme in detail. The result also shows that several ablation studies strengthen our proposed innovations. These experiments indicate that unlabeled data has a significant contribution to data augmentation using the conditional semi-supervised scheme for spam detection.

**Keywords**

**Spam detection; text classification; deep generative model; data augmentation; semi-supervised learning.**

## 1. Introduction

The growth of deep learning has penetrated various field studies. In recent years, the contribution of deep learning has presented a significant impact resulting in state-of-the-art image processing and natural language processing (NLP) (Minaee et al., 2022; Khurana et al., 2023). The development of NLP specifically has undergone very rapid progress. NLP has various primary tasks such as text classification, natural language inference, text generation, and others (Liu et al., 2021; Khan et al., 2023). Then, one of the significant tasks in NLP applications is spam detection or classification. Spam is a message having the aim of any prohibited aim such as fraud, fake news, promotion, and others (Jáñez-Martino et al., 2023; Jain, Yadav, & Choudhary, 2020). Many spam messages often come to us through email or mobile messages. Based on the DataProt reports, 56.5% of incoming emails in 2022 were categorized as spam (Cveticanin, 2023). Therefore, unwanted messages such as spam or scam are critical to be filtered.

Various studies related to spam detection have been conducted by researchers using NLP approaches. One of the crucial parts of NLP to perform various tasks lies in text representation and end-to-end training (Li, 2018). Text representation makes the computer understand text-based data by representing them in numerical form. Whereas end-to-end training is a deep learning technique where the model learns and trains all the steps simultaneously. A work of spam detection conducted by Galeano (2021) on short messages exploited several text representation schemes such as bidirectional encoder representations from transformers (BERT) or Bag-of-Words (BoW) and various classifiers for final classification. Then, the result showed that the best text representation is BERT outperforming other text representation models. Ahmed et al. (2022) performed a survey to find the best machine learning model for classifying spam emails. They revealed that excellent machine learning algorithms to filter those emails are support vector machine (SVM), Naïve Bayes, and logistic regression. Another spam detection scheme introduced by Yerima and Bashar (2022) proposed semi-supervised spam message classification using one-class SVM (OC-SVM) that only trained non-spam messages. Their proposed scheme outperformed the supervised learning with vectorization methods of text representation.

However, there are several challenges encountered by researchers in the spam detection task. The availability of data is one of them. Most schemes that performed the spam detection tasks utilized supervised learning for their approaches (Ahmed et al., 2022). Whereas supervised learning necessitates extensive labeled data to attain a robust model (Aljuaid & Anwar, 2022). On the other hand, the availability of labeled data becomes one issue, especially in other than English such as the Indonesian language. Finding large-scale labeled data will be expensive and time-consuming. Therefore, we introduce data augmentation to overcome those mentioned issues. To extend the training data, we then exploit unlabeled data that are easier to obtain.

Finally, this paper proposes conditional semi-supervised data augmentation (CSSDA) for spam detection. We will address the issue of the lack of availability of labeled data in spam messages. The main architecture of our proposed model is a deep generative model based on adversarial learning. Here, our proposed model results in latent variables as fake samples from unlabeled data. While real latent variables come from labeled data representation features combined with its label embedding. Then, these latent variables become the input for the final classifier. We introduce latent variables rather than text representation features as generated data since unlabeled and labeled data possess a similar distribution, the difference is only that the labeled data have extra information related to their annotation (Yu et al., 2020). In addition, CSSDA performs the conditional generative model to create the latent variable influenced by the label embedding. Thus, the resulting synthetic data is not divergent from real data characteristics. Then, a discriminator acting as the classifier will identify whether the input is real or fake samples. Here, the discriminator also categorizes input samples into the related class if the input is real latent variables. To minimize vanishing gradient and bias-shift error, we perform derivation of the unsupervised loss during the training process explained in the third section.

Based on the above analysis, the major contributions of our study are listed as follows:

- This study proposes a detection model for spam messages that lack the availability of labeled data.
- Our main framework is a conditional semi-supervised deep generative model exploiting unlabeled data to extend training data.
- We introduce the latent variable as generated synthetic data to implement conditional generation.
- We report the result comparison between our proposed model and several counterpart models in performing spam detection in Indonesian Short Message Service (SMS) datasets.

The rest of this paper is organized as follows. In Section 2, we present several related works. We illustrate the basic theory and framework of our proposed model in Section 3, and our proposed model will be presented in detail. Section 4 presents the experimental results and analysis of the results. Finally, Section 5 concludes this study.

## 2. Related work

There are several proposed schemes to overcome the lack of labeled data. One of them is data augmentation, a scheme utilizing a particular algorithm to generate new synthetic data from the available data (Shorten, Khoshgoftaar, & Furht, 2021). However, the implementation of data augmentation in NLP is more complicated than in image processing. Since the image originally is composed of individual pixels represented in numeric form, the data augmentation can be performed by adding some noises, translation, color space transformation, and others without losing the critical information. In discrete data such as text processing, the data augmentation encounters difficulties during the sampling process (Wu & Wang, 2020). Alkadri, Elkorany, and Ahmed (2022) conducted data augmentation for the labeled spam in Arabic tweets. Their scheme randomly substitutes several words into their synonyms based on similarity. They performed the replacement of word tokens based on the similarity score also considering the surrounding words. They changed about 60% of word tokens from the original data to generate new sentences. However, the data augmentation by synonym replacement has lacked generating new linguistic features, because their sentence structure is the same between the synthetic and original data. Another scheme to overcome the lack of labeled data is unsupervised data augmentation (UDA) based on consistency training. Xie et al. (2020) introduced UDA to address the lack of labeled data exploiting unlabeled data. The use of unlabeled data is for performing unsupervised consistency training. They augmented unlabeled data through simple-noising operations such as back-translation. Their proposed scheme tries to minimize unsupervised consistency training loss between unlabeled data and noised unlabeled data.

Another of the influential state-of-the-art works to augment data is generative adversarial networks (GAN) (Tran et al., 2021). The GAN model has two primary components, they are generator and discriminator which perform adversarial learning. The generator will generate some synthetic data based on real data distribution, whereas the discriminator distinguishes whether the coming input is real samples or fake samples (Fan et al., 2022). The generator will try to fool the discriminator with its generated samples. Then, the discriminator should be able to detect that the generated sample from the generator is fake. A standard GAN in an image processing task can be extended by adding some extra information in the generator or the discriminator as a conditional scheme (Jin, Ye, & Li, 2023; Mirza & Osindero, 2014). The extra information added to both components can be data labels or other modalities. For example, the noise vector and the extra information are utilized as the generator inputs. Their experiment in applying the conditional GAN in image processing tasks showed promising results. However, the GAN model is quite complicated to be implemented in text-based data since GAN is only introduced to result in real-numeric values, especially in image processing (De Rosa & Papa, 2022; Yu et al., 2017). The generative model in text processing has a main challenge in the discrete state space issue (Alsmadi et al., 2022).

Croce, Castellucci, and R. Basili (2020) proposed GAN-BERT performing text classification tasks in few labeled data. Their model exploits an improved GAN scheme proposed by Salimans et al. (2016), implemented originally for image-processing tasks. In this improved GAN model, the role of the discriminator does not only distinguish whether the inputs are real or fake samples but also classifies them into related classes if the inputs originally are labeled data. Then, GAN-BERT brought this improved GAN model into text classification by adding BERT architecture to extract text representation features. Although GAN-BERT performance outperforms the standard BERT model, the performance immediately will be overcome by the standard BERT model if the availability of the labeled data increases. Then, Riyadh and Shafiq (2022) proposed GAN-BElectra extending the GAN-BERT architecture to give pseudo labels of the unlabeled data. Then, labeled data and unlabeled data that already received the pseudo labels were trained using an Electra-based pre-trained model. However, the performance of text classification using GAN-BElectra was not satisfied. Based on these previous works, existing machine learning models perform spam detection that relies on huge training data. Other methods that perform data augmentation did not utilize unlabeled data more advanced.

## 3. Methods

In this section which first introduces the research objectives, we describe the basic method applied in conditional semi-supervised data augmentation using deep generative model for text classification tasks. In addition, the training process of our proposed model is discussed.

### 3.1. Research objective

As discussed in Section 2.2, text classification including spam detection using machine learning has some challenges since this task requires a large amount of labeled data, that is frequently difficult and costly to acquire. By investigating the previous studies, we reveal that data augmentation is a precise approach to address this issue by expanding training samples. However, applying data augmentation in the machine learning model does not guarantee improving the model's performance (Shi et al., 2024). Consequently, our first research objective is to develop a robust scheme for scientific text classification of spam messages with limited labeled data. Additionally, to enhance the model's spam detection, our second research objective focuses on how the proposed model can utilize unlabeled data. We define the conditional scheme by forming latent variables as generated data obtained from the generator and label embedding through element-wise operation. This approach enables the model to possess more samples utilized for the training process of spam detection. By introducing this two-stance information, this paper aims to obtain a robust model of spam message detection.

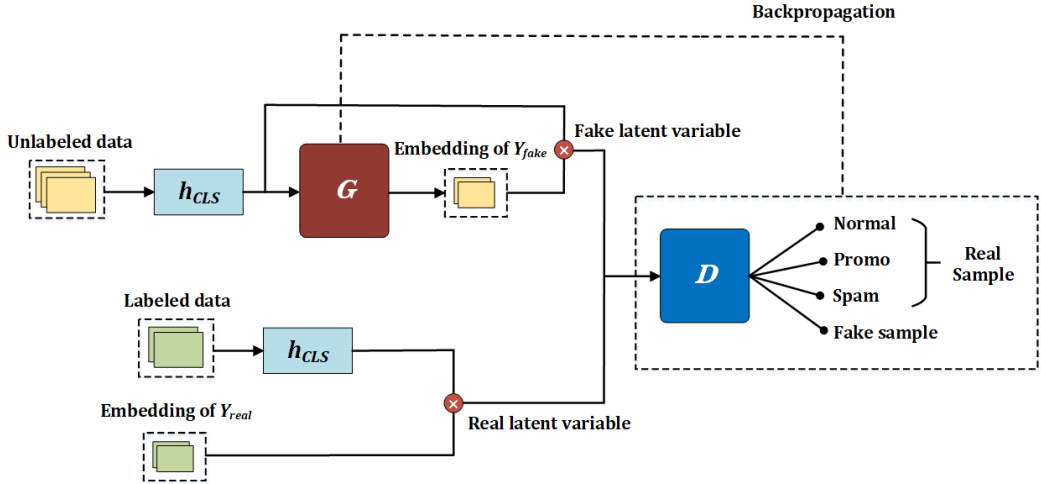

**Fig. 1**. The architecture of our proposed model consisting of the generator *G* and the discriminator *D*.

### 3.2. Framework of the proposed model

#### 3.2.1. Our proposed model architecture

The architecture of our proposed model is shown in Fig. 1. Here, our proposed CSSDA exploits both unlabeled data and labeled data to perform semi-supervised data augmentation for spam message detection. Before we bring text data to the generator or discriminator, we need to transform the message into a vector $h_{CLS}$ of sentence-level representation. Sentence-level representation is the numerical feature vector of the message texts so that the computer can understand the text data. In our proposed model, we utilize transformer-based encoder, i.e., BERT and its variant, as the text representation model. BERT is a pre-trained model that learns linguistic information from the corpus during the pre-training process using deep learning schemes. Given a sentence $t = (w_1, ..., w_n)$, BERT will generate an output in $n + 2$ vector representations, i.e., $(h_{CLS}, h_{w1}, ..., h_{wn}, h_{SEP})$.

Then, the sentence-level representation of the unlabeled message will be input for the generator to produce the fake label $y_{fake}$. Our proposed CSSDA employs the generator to produce an embedding of fake label $y_{fake}$, instead of fake sentence representation features. The embedding of fake label $y_{fake}$ from the generator is then combined with the unlabeled message representation vector $h_{CLS}$ by performing element-wise operation to generate the fake latent variable $v_{fake}$. On the other hand, the real latent variable $v_{real}$ is obtained by combining the sentence-level representation vector $h_{CLS}$ of the labeled message with its real label $y_{real}$ through element-wise operation. The purpose of producing the latent variable is to perform a conditional generative model. Thus, the distribution pattern of latent variables is similar to the distribution pattern of real message labels. Then, the discriminator will attempt to identify the fake latent variable and the real latent variable. If the input sample is detected as the fake latent variable, the discriminator should classify it into the $k + 1$ label. In addition, the discriminator should classify the input sample into one of the $k$ classes when the coming input is the real latent variable. The training process of our proposed CSSDA is summarized in Algorithm 1.

**Algorithm 1** The training process of CSSDA.
_______________________________________________
**Input**: Labeled message set $L_m$, Unlabeled message set $U_m$,
$Y_l$ = Label set of $L_m$, All message set $T = \{L_m, U_m\}$
**Output**: Trained model
1: **Begin**
2:   **for** $t \in T$ **do**
3:     **if** $t \in U_m$ **then**
4:       $h_{CLS}$ = TextRepresentation($t$)
5:       $y_{fake}$ = $G(h_{CLS})$
6:       $v_{fake}$ = $h_{CLS} \otimes y_{fake}$
7:     **else if** $t \in L_m$ **then**
8:       $h_{CLS}$ = TextRepresentation($t$)
9:       $v_{real}$ = $h_{CLS} \otimes y_l$    where $y_l \in Y_l$
10:     **end if**
11:   **end for**
12:   $V = \{V_{fake}, V_{real}\}$
13:   **for** $v \in V$ **do**
14:     **if** $v \in V_{fake}$ **then**
15:       assign the final class: $k + 1$ label $\leftarrow D(v)$
16:     **else if** $v \in V_{real}$ **then**
17:       assign the final class: one of $k$ labels $\leftarrow D(v)$
18:     **end if**
19:   **end for**
20: **End**
_______________________________________________

### 3.2.2. Text data representation

We select the Transformer-based model as the representation model of the texts to obtain the feature vector such as BERT and its variant. The original model of BERT is a multi-layer stack of Transformer encoders. The model utilizes the encoders to learn the sequence information, bidirectionally. The BERT model attains state-of-the-art in diverse NLP tasks since it proposed two schemes during the pre-training process. They are masked language modeling (MLM) mechanism and next sentence prediction (NSP) mechanism.

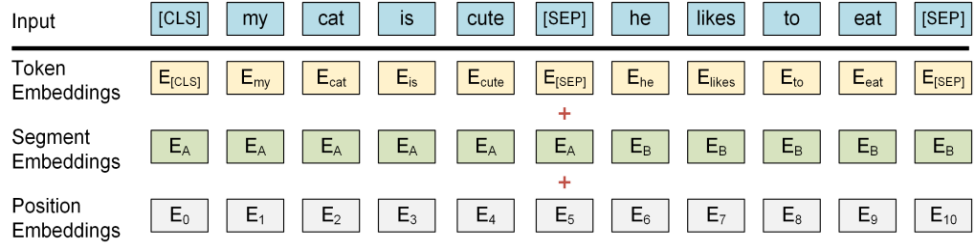

**Fig. 2.** Input representation scheme of the BERT model, adopted from (Devlin et al., 2019).

Input representation of the BERT model and its variant such as RoBERTa (Robustly Optimized BERT Pretraining Approach) is composed of three layers, they are token embeddings, segment embeddings, and position embeddings as shown in Fig. 2. Token embeddings are subword tokens of the input sequence gained by the WordPiece algorithm, they always start with a special token [CLS] for sentence-level representation for classification and end with a special token [SEP] as a separator token. Segment embeddings denote the position of a specific token belonging to either sentence *A* or sentence *B*. Position embedding represents the position of each word token in the sequence. All these embeddings are then summed up with an element-wise operation.

### 3.2.3. Conditional generative model

The generator architecture basically is a multi-layer perceptron (MLP) with the input coming from the vector $h_{CLS}$ of the representation of unlabeled messages. In particular, the multi-layer perceptron in the generator is one hidden layer with a leaky rectified linear unit (ReLU) as the activation function. Then, a dropout is applied after this hidden layer to minimize overfitting. The vector dimension of the embedding of fake label $y_{fake}$ is 768- dimension such as BERT output dimension size. As shown in Fig. 1, we multiply the embedding of the fake label $y_{fake}$ and the representation vector of the unlabeled message $h_{CLS}$ with the element-wise operation to generate fake latent variable $v_{fake}$ as conditional data generation.

Afterward, the discriminator is basically also the multi-layer perceptron that consists of one hidden layer with the leaky-ReLU as the activation function and a softmax layer as the final layer. The discriminator inputs are the fake latent variable $v_{fake}$ and the real latent variable $v_{real}$. The real latent variable $v_{real}$ is obtained by the representation vector of the labeled message $h_{CLS}$ multiplied with the embedding of real label $y_{real}$ in the element-wise operation. The use of the real label combined with the vector representation is to control the model when generating fake data. Thus, the generated fake data have an imminent similarity to the real data distribution through adversarial learning. In the testing process, the CSSDA model relegates $1 + k$ label as the final class.

### 3.2.4. The proposed semi-supervised learning

Let us assume, $p_d$ and $p_g$ respectively represent the real latent variable distribution and the generated or fake latent variable distribution. We define $p_m(\bar{y} = y | v, y = k+1)$ as the probability from the model $m$ where the latent variable $v$ is attributed to the fake class. While $p_m(\bar{y} = y | v, y = 1,...,k)$ is the probability that the latent variable $v$ is associated with real data belonging to the one of $k$ real classes. In the relationship between the generator and the discriminator, there is a competition called adversarial learning in which the generator tries to deceive the discriminator, and the discriminator attempts to distinguish whether the input data are real or not. To optimize adversarial learning, the lost function comes from the generator and the discriminator. Since the proposed model in this study performs semi-supervised learning, the loss function comprises supervised loss and unsupervised loss. Then, we formulate the discriminator loss $L_D$ as a summation of the discriminator's supervised loss $L_{Dsup}$ and the discriminator's unsupervised loss $L_{Dunsup}$. The discriminator's supervised loss calculates the prediction error of the model in classifying the class of labeled data. The expected value (E) of discriminator's supervised loss $L_{Dsup}$ can be formulated as follows:

$$L_{D\sup} = -\mathrm{E}_{v,y \sim p_d}\left[\log p_m(\bar{y} = y | v, y \in \{1,...,k\})\right], \tag{1}$$

where $v$ is the input for the discriminator assumed from the real latent variable. Then, the discriminator as the classifier will output a class among the $k$ classes. Since the original output is in the form of a $k$-dimensional vector of logits, we apply the cross-entropy loss function between the actual label $y$ and the predictive distribution $(\bar{y} = y | v)$.

Then, the discriminator's unsupervised loss $L_{Dunsup}$ calculates the error in incorrectly detecting the real latent variable as fake and not recognizing the fake latent variable. The discriminator's unsupervised loss $L_{Dunsup}$ can be formulated with the following equation:

$$L_{Dun\sup} = -\mathrm{E}_{v \sim pd}\left[\log(1 - p_m(\bar{y} = y | v, y = k+1))\right] - \mathrm{E}_{v \sim p_g}\left[\log p_m(\bar{y} = y | v, y = k+1)\right], \tag{2}$$

$$L_{Dun\sup} = -\mathrm{E}_{v \sim pd} \log D(v_{real}) - \mathrm{E}_{v \sim pg} \log(1 - D(v_{fake})), \tag{3}$$

where we substitute $1-p_m(y=k+1)$ in Eq. (2) to $D(v)$ in Eq. (3). $D(v)$ is the softmax function operation over the $k$-dimensional logits and a vacant logit $l_{k+1}(v)$ of the fake class. Then, we set the logit of the fake class $l_{k+1}(v)$ to 0 (zero) resulting in $e^0=1$ since this vacant logit has no impact to the output of $k$ classes (Salimans et al., 2016). The discriminator $D(v)$ can be represented as follows:

$$D(v) = \frac{Z(v)}{Z(v)+1}, \qquad (4)$$

$$Z(v) = \sum_i^k e^{l_i(v)}. \qquad (5)$$

Based on the equations above, Eq. (3) can be substituted through Eqs. (4) and (5) as below:

$$L_{Dun\sup} = -\mathrm{E}_{v \sim pd} \log\left(\frac{Z(v_{real})}{Z(v_{real})+1}\right) - \mathrm{E}_{v \sim pg} \log\left(1 - \frac{Z(v_{fake})}{Z(v_{fake})+1}\right), \qquad (6)$$

$$L_{Dun\sup} = -\mathrm{E}_{v \sim pd} \log Z(v_{real}) + \mathrm{E}_{v \sim pd} \log(Z(v_{real})+1) + \mathrm{E}_{v \sim pg} \log(Z(v_{fake})+1), \qquad (7)$$

$$L_{Dun\sup} = -\mathrm{E}_{v \sim pd} \log \sum_i^k e^{l_i(v_{real})} + \mathrm{E}_{v \sim pd} \log(\sum_i^k e^{l_i(v_{real})}+1) + \mathrm{E}_{v \sim pg} \log(\sum_i^k e^{l_i(v_{fake})}+1). \qquad (8)$$

Then, we introduce the Softplus function, a soft approximation of the ReLU function, to Eq. (8). In the Softplus function, the output will always be in the form of positive values. Since the Softplus function can be distinguished in a range of ($-\infty$, $+\infty$), it can perform backpropagation of the gradient across the range (Zhao et al., 2017). Thus, the Softplus function generates close real characteristics and can control vanishing gradient and bias-shift error. Since the Softplus function is formulated in $Softplus(x) = \log(\exp(x) + 1)$, Eq. (8) can be expressed in Eq. (9) as following:

$$L_{Dun\sup} = -\mathrm{E}_{v \sim pd} \log \sum_i^k e^{l_i(v_{real})} + \mathrm{E}_{v \sim pd} Softplus(\sum_i^k e^{l_i(v_{real})}) + \mathrm{E}_{v \sim pg} Softplus(\sum_i^k e^{l_i(v_{fake})}). \qquad (9)$$

Our proposed model also formulates the generator loss $L_G$ as the summation of the generator's feature matching loss $L_{Gfm}$ and the generator's unsupervised loss $L_{Gunsup}$. The generator's feature matching loss $L_{Gfm}$ is presented to calculate how the generator forms latent variables approximating the statistics of real latent variables. In other words, the generator has to attempt to form fake samples as similar as possible to the expected value (E) of characteristics of the real latent features in the discriminator intermediate layer. Let's define $f(v)$ as the activation of the discriminator intermediate layer, the formula for the generator's feature matching loss $L_{Gfm}$ is expressed below:

$$L_{Gfm} = criterionG\left(\mathrm{E}_{v \sim p_d} f(v_{real}), \mathrm{E}_{v \sim p_g} f(v_{fake})\right), \qquad (10)$$

where we define *criterionG* as a parameter to calculate the generator's feature matching loss. We practically utilize mean squared error (MSE) as *criterionG*. Since there is adversarial learning between the generator and the discriminator through a backpropagation scheme, our proposed model also considers the unsupervised loss of the generator. The generator's unsupervised loss $L_{Gunsup}$ computes the error of the fake latent variable that correctly is identified by the discriminator. This loss function has a similar process to the unsupervised loss of the discriminator but ignores the error in detecting the real latent variable. The generator's unsupervised loss $L_{Gunsup}$ can be formulated

in Eqs. (11-15) below:

$$L_{Gun\sup} = -E_{v \sim p_g}\left[\log(1 - p_m(\bar{y} = y | v, y = k + 1))\right], \quad (11)$$

$$L_{Gun\sup} = -E_{v \sim p_g} \log D(v_{fake}), \quad (12)$$

$$L_{Gun\sup} = -E_{v \sim p_g} \log\left(\frac{Z(v_{fake})}{Z(v_{fake}) + 1}\right), \quad (13)$$

$$L_{Gun\sup} = -E_{v \sim p_g} \log Z(v_{fake}) + E_{v \sim p_g} \log(Z(v_{fake}) + 1), \quad (14)$$

$$L_{Gun\sup} = -E_{v \sim p_g} \log \sum_i^k e^{l_i(v_{fake})} + E_{v \sim p_g} Softplus(\sum_i^k e^{l_i(v_{fake})}). \quad (15)$$

## 4. Experimental evaluation and discussion

In this section, we will mainly evaluate the performance of our proposed model, i.e., CSSDA, and compare it with several counterpart models. In addition, this section also conducts several ablation studies to identify the effect of each component on CSSDA performance.

### 4.1. Datasets and setup

#### 4.1.1. Datasets

We take Indonesian SMS datasets to build spam detection from Rahmi and Wibisono (2016) and Tandra et al. (2021). We select Indonesian SMS texts as the dataset since the labeled data in this language lacks the availability. The labeled SMS datasets consist of three labels, they are spam, promo, and normal messages. Spam messages contain fraudulent information such as unknown bills and irrational gifts. Promo messages are product, business, or service advertisements. Then, normal messages are common chats with friends or colleagues, in other words besides promos and spam. We set the Rahmi and Wibisono dataset containing 335 spam messages, 239 promo messages, and 569 normal messages as the training data and the Tandra dataset containing 578 spam messages, 1957 promo messages, and 811 normal messages as the testing data. In the Tandra dataset, there are one-time password (OTP) labeled messages that we assign them as normal messages. The detail of the percentage of each class data in our used dataset is depicted in Fig. 3. Then, we apply duplicate data removal for samples in training data that are duplicated in testing data. Thus, the Rahmi and Wibisono dataset as training data consists of 2243 samples with 269 normal, 253 spam, and 1721 promo messages. The examples of SMS datasets are shown in Table I.

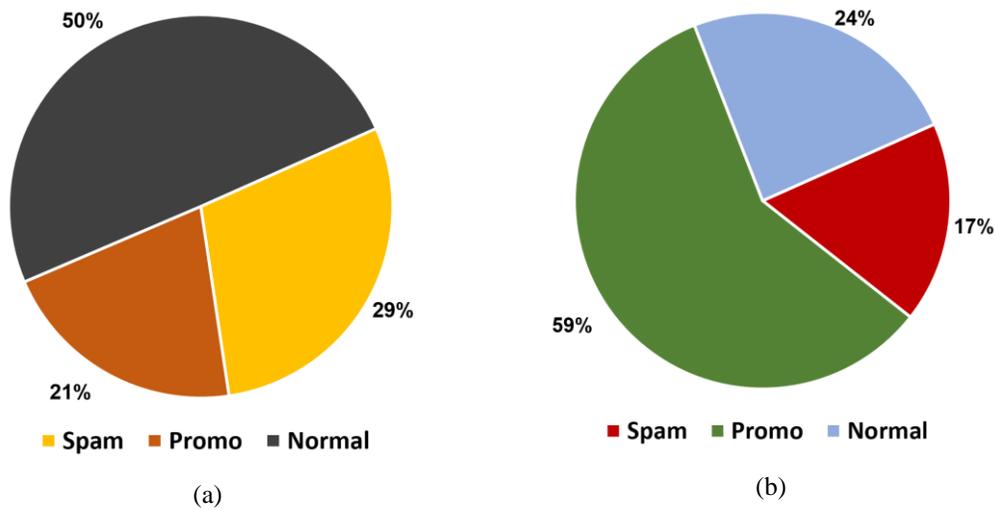

**Fig. 3.** (a) The percentage of each data class in the Rahmi and Wibisono dataset. (b) The percentage of each data class the Tandra dataset.

| SMS text | Category |
|---|---|
| Original texts: Info resmi tri care selamat nomor anda terpilih mendapatkan hadiah 1 unit mobil dri tri care dengan PIN Pemenang: br25h99 info: www.gebeyar-3care.tk | Spam |
| English version: An official info from Tri care, congratulations, your number was selected to win a prize of 1 unit of a car from Tri care with Winning PIN: br25h99 info: www.gebeyar-3care.tk | Spam |
| Original texts: Gunakan GrabCar atau GrabTaxi & dapatkan diskon 35RIBU ke dan dari Stasiun/Bandara di Bandung. Masukkan kode promo GRABTRAVEL hingga pemakaian 30 September 2016. Hatur Nuhun. | Promo |
| English version: Use GrabCar or GrabTaxi & get a 35K discount to and from stations/airports in Bandung. Enter the promo code GRABTRAVEL until 30 September 2016. Thank you. | Promo |
| Original texts: Coba siapa yg lagi di prodi? Punten liatin jadwal sidang. Ada tulisan suruh kumpul jam brp gitu gak? Nuhun. | Normal |
| English version: Who is currently in the department office? Please look at the defense schedule. Is there any instruction on what time to gather or not? Thank you. | Normal |
| English version: To show your ringtone with "heart voice", a hit from Ayu Tingting, reply to this SMS, type yes, 808 | Normal |

**Table 1. Examples of message texts.**

#### 4.1.2. Experimental setup

For the text representation model, we exploit the pre-trained Transformer-based model with 768-dimensional embeddings. Then, we set the batch size to 64 and the epoch size to 5 during the training process since the number of data training is quite a lot. As the labeled SMS datasets have three categories, i.e., spam, promo, and ham (normal), we set the $k$ value = 3 (three) for the discriminator parameter. In the training process, we perform three training schemes to evaluate the robustness and consistency of the model in various amounts of unlabeled data. For the first scheme, we set training data in which 0.25 of training data is labeled data and the rest is unlabeled data. Then, the second scheme utilizes the unlabeled data same as the number of labeled data for the training process with the ratio of 0.5:0.5 of training data. Lastly, we set training data for the third scheme in which 0.75 of training data is labeled data and the rest is unlabeled data.

### 4.2. Evaluation metrics

Since the amount of data in each class of datasets is not balanced, we utilized precision, recall, balanced accuracy, and F-score as evaluation metrics that are significant rather than standard accuracy (Bej et al., 2021; Ding et al., 2023). Precision can be regarded as a measure of the model to identify only relevant data points, while recall is a measure of the model to identify all relevant cases within the dataset. Balanced accuracy is the average of the correct prediction rates for each class separately. Then, the F-score is a harmonic mean score between precision and recall values. The formulas of the evaluation metrics are as follows:

$$precision = \frac{TP}{TP + FP}, \qquad (16)$$

$$recall = \frac{TP}{TP + FN}, \qquad (17)$$

$$balanced\_accuracy = \frac{1}{2}\left(\frac{TP}{TP + FN} + \frac{TN}{TN + FP}\right), \qquad (18)$$

$$F-score = \frac{2 \times precision \times recall}{precision + recall}. \qquad (19)$$

where *TP* (true positive) is a scenario in which the model accurately identifies the positive class. *FP* (false positive) is a condition in which the model incorrectly identifies the positive class. *FN* (false negative) occurs when the model incorrectly identifies the negative class. Then, *FN* (false negative) is the result when the model incorrectly identifies the negative class.

### 4.3. Performance comparison

To assess the performance of our proposed model in spam detection with the lack of availability of labeled data, we conduct the performance comparison with several counterpart models. We present the performance comparison in several state-of-the-art algorithms using both machine learning and deep learning. In the machine learning approach, we take several algorithms presented by Abid et al. (2022). They are SVM, random forest, and logistic regression as the classifier while using the term frequency-inverse document frequency (TF-IDF) model as the text representation that is based on the vectorization method. However, the machine learning approach only exploits the labeled data since it does not perform data augmentation. The results of machine learning models for spam detection are presented in Table 2 in which we set all training data as labeled data. The results show that machine learning models underperform in message classification since there is no model achieving a metric score of more than 60%. We argue these machine learning models cannot address the prediction on testing data with large samples.

| Method | Balanced accuracy | Precision | Recall | F-score |
|---|---|---|---|---|
| SVM | 39.83 | 52.56 | 42.30 | 39.73 |
| Random forest | 42.28 | 58.40 | 51.14 | 44.78 |
| Logistic regression | 33.72 | 38.50 | 22.77 | 12.38 |

\* All values are in the percentage form

**Table 2. Result comparison of machine learning schemes.**

For the deep learning approach, the counterpart models are GAN-BERT, UDA, and GAN-BElectra. These deep learning algorithms perform data augmentation to improve the task of spam detection. Therefore, these schemes utilize unlabeled data of messages. For the UDA model, back translation is performed as the augmentation scheme. Tables 3, 4, and 5 present the performance result of spam detection based on the deep learning approach and data augmentation including our proposed model, i.e., CSSDA. Table 3 shows the result comparison of our model with the counterpart models in which 25% of training data is labeled data and the rest is unlabeled data. Table 4 shows the result comparison of performance in which labeled data and unlabeled data have the same amount. Then, Table 5 presents the result comparison of our model with the counterpart models in which 75% of training data is labeled data and the rest is unlabeled data. All metric values in all tables are in percentage values.

| Method | Balanced accuracy | Precision | Recall | F-score |
|---|---|---|---|---|
| GAN-BERT | 36.15 | 55.82 | 33.54 | 21.50 |
| UDA | 62.63 | 80.89 | 63.05 | 59.64 |
| GAN-BElectra | 66.28 | 72.67 | 59.55 | 59.65 |
| Our proposed model | **76.48** | **79.26** | **70.84** | **70.58** |

* All values are in the percentage form

**Table 3. Result comparison of deep learning exploiting less labeled data.**

| Method | Balanced accuracy | Precision | Recall | F-score |
|---|---|---|---|---|
| GAN-BERT | 33.33 | 8.61 | 29.33 | 13.31 |
| UDA | 69.43 | 82.10 | 71.37 | 67.83 |
| GAN-BElectra | 66.28 | 72.67 | 59.55 | 59.64 |
| Our proposed model | **84.82** | **85.70** | **85.46** | **85.51** |

* All values are in the percentage form

**Table 4. Result comparison of deep learning exploiting the amount of unlabeled data same as labeled data.**

| Method | Balanced accuracy | Precision | Recall | F-score |
|---|---|---|---|---|
| GAN-BERT | 33.65 | 11.41 | 29.51 | 16.42 |
| UDA | 75.08 | 83.66 | 71.10 | 72.92 |
| GAN-BElectra | 78.21 | 80.80 | 77.58 | 78.19 |
| Our proposed model | **85.23** | **86.10** | **83.10** | **83.40** |

* All values are in the percentage form

**Table 5. Result comparison of deep learning exploiting much labeled data.**

The results above indicate that our proposed model generally outperforms the others in terms of balanced accuracy, precision, recall, and F-score metrics. In Table 3, we can see that our proposed model achieves robust performance obtaining the highest values in all evaluation metrics with values more than 70% in which we exploit few labeled data. Based on the score of the balanced accuracy metric, our proposed model can correctly identify 76.48% of testing samples by precisely detecting the samples more than 79.26% of the time. Referring to the results of the recall metric, our proposed CSSDA correctly predicts 70.84% of samples that belong to the target class out of the total samples for that class. Table 4 shows the results in which we exploit the same amount of unlabeled data and labeled data, yet our proposed model still generally outperforms the others achieving the best values in all metrics. In this experiment scheme, CSSDA stands out as the model consistently achieving the balanced accuracy of 84.82% and the F-score of 85.51%. When we proposed a scheme in which labeled data in the training process is greater than unlabeled data, the result of evaluation metrics in Table 5 shows that CSSDA still outperforms other counterpart models by obtaining the balanced accuracy and precision metrics with values of more than 85%. These three experiment schemes in Tables 3, 4, and 5 denote that the proposed CSSDA is superior to others with a major gap in evaluation metric scores compared to counterpart models. The proposed CSSDA obtains high and robust accuracy values when using more label data denoted by results in Tables 4 and 5.

In all experiments, we can see that the GAN-BERT model seems underperforming. Its all-metrics scores cannot exceed more than 60%. We argue that the GAN-BERT model needs more labeled data in the training process. In the original paper of GAN-BERT, the performance in the sentiment classification task also showed that the result was poor. GAN-BElectra also fails to obtain excellent performance since this model depends on the GAN-BERT model. Then, the results of the UDA model cannot equalize our proposed model performance. The UDA algorithm utilizes unlabeled data only to improve consistency training through noise injection. On the other hand, our proposed CSSDA performs consistency training through $k + 1$ class detection of the fake latent variable. Besides, consistency training is also performed through the real latent variable detection when it is correctly identified. Thus, our unsupervised loss in performing consistency training consists of two components, they are not recognizing the real data and incorrectly detecting the fake data. Moreover, the model controls the quality of fake data through conditional generative adversarial networks. We also present Fig. 4 to show the receiver operator characteristic (ROC) curve of the proposed CSSDA to illustrate the trade-off between true and false positives in the experiment in which the amount of unlabeled data and labeled data is the same. Since we apply the ROC curve to multi-class classification, we designate one class as the positive class while treating all other classes as the negative class. Fig. 4. shows that the area under the ROC curve of each class exceeds 0.85 and, moreover, the normal and promo channels can reach a score of 0.90. This result indicates that the proposed CSSDA can distinguish between message classes, excellently.

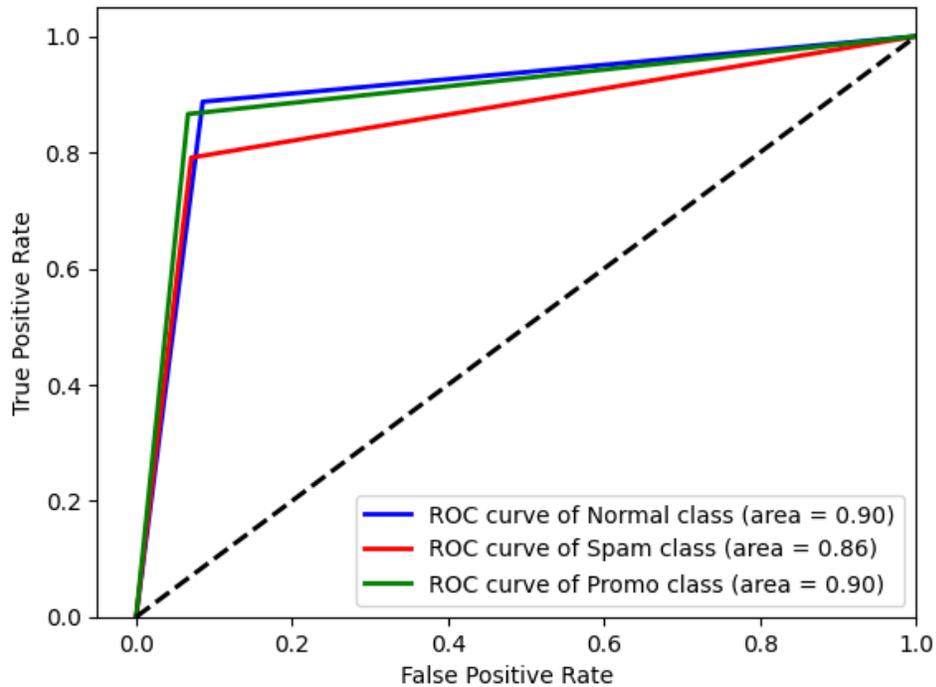

**Fig. 4.** The receiver operator characteristic (ROC) curve of the proposed CSSDA.

Since all the algorithms of message classification in the previous experiments use the pre-trained BERT model for text representation, we also investigate the results when using the pre-trained RoBERTa model, a variant of BERT, for text representation. We utilize the pre-trained RoBERTa model available from HuggingFace[1]. Table 6 presents the results of this experiment, conducted with unlabeled data and labeled data having the same ratio.

| Method | Balanced accuracy | Precision | Recall | F-score |
| --- | --- | --- | --- | --- |
| GAN-RoBERTa | 33.65 | 11.41 | 29.51 | 16.42 |
| UDA | 58.81 | 70.31 | 51.14 | 50.89 |
| GAN-BElectra | **66.63** | 73.01 | 61.03 | 61.69 |
| Our proposed model | 65.78 | **75.73** | **62.52** | **62.85** |

* All values are in the percentage form

**Table 6. Result comparison on data augmentation approach.**

Even with the text representation model changed to RoBERTa, the proposed model continues to achieve satisfactory performance compared to the other models. CSSDA attains the highest values in precision, recall, and F-score. The proposed only fails to obtain the highest value in balanced accuracy outperformed by BERT-Electra with a slight difference.

---

[1] https://huggingface.co/cahya/roberta-base-indonesian-522M

### 4.4. Ablation study on data augmentation

| Method | Balanced accuracy | Precision | Recall | F-score |
|---|---|---|---|---|
| Our proposed model | **84.82** | **85.70** | **85.46** | **85.51** |
| Non-data augmentation | 81.39 | 83.48 | 80.56 | 80.78 |

\* All values are in the percentage form

**Table 7. Result comparison on data augmentation approach.**

Since our work proposes the deep learning model along with data augmentation, the proposed model also exploits the unlabeled data. In this ablation study, we will assess how the deep learning model performs the task of spam detection but without performing data augmentation. Therefore, we utilized the BERT-MLP model for the counterpart approach as also proposed for spam detection in Fahfouh et al. (2022). The model utilizes the BERT model as text representation and MLP as the classifier. The result will answer whether the data augmentation can improve the performance of spam detection or not. Table 7 shows the result comparison of this ablation study between the standard BERT-MLP model without data augmentation and our proposed model that exploits all unlabeled data for data augmentation. We set half of training data in labeled data and the rest is unlabeled data. We can see that the non-data augmentation approach of the BERT-MLP model fails to outperform our proposed model in all evaluation metrics compared. This indicates that the data augmentation approach in our proposed model can enhance the performance of spam detection.

### 4.5. Ablation study on conditional generative model

| Method | Balanced accuracy | Precision | Recall | F-score |
|---|---|---|---|---|
| Our proposed model | **84.82** | **85.70** | **85.46** | **85.51** |
| Non-conditional generation | 84.05 | 84.16 | 81.35 | 81.63 |

\* All values are in the percentage form

**Table 8. Result comparison of conditional generative model.**

The generative scheme in our proposed model aims to form the fake latent variable instead of fake text representation features. Therefore, CSSDA can control latent variable features through element-wise multiplication between label embedding and message representation feature vectors. We compare the performance of the conditional generative model with the generator that only produces text representation features with noise as the input of the generator. Here, we set half of training data to labeled data and the rest is unlabeled data to perform this ablation study. In this experiment, we set our proposed model with the conditional generative adversarial network, and without it.

Table 8 shows the result comparison of the ablation study on the conditional generative model. The result indicates that the proposed model exploiting the conditional scheme can improve performance. Through the conditional generative, generated data characteristics are forced to have similarity to real data characteristics. This paper performs the conditional generative scheme by forming the fake latent variable from the combination of sentence representation vector $h_{CLS}$ with the fake label from the generator.

### 4.6. Ablation study on loss derivation

| Method | Balanced accuracy | Precision | Recall | F-score |
|---|---|---|---|---|
| Our proposed model | **84.82** | **85.70** | **85.46** | **85.51** |
| Non-loss derivation | 59.58 | 60.99 | 57.79 | 58.28 |

* All values are in a percentage form

**Table 9. Result comparison of proposed loss derivation.**

In our proposed CSSDA, we perform substituting and derivation schemes for unsupervised loss both for the generator and the discriminator. This ablation section presents the counterpart model that does not substitute and derives Eq. (2) from the discriminator's unsupervised loss and Eq. (11) from the generator's unsupervised loss. Our proposed model implementing loss derivation achieves the best result by achieving scores of about 85% in all metrics as shown in Table 9. When our proposed model does not implement loss derivation in Eq. (2) and Eq. (11), the performance result of evaluation metrics cannot even exceed values of 70%. The findings from this ablation study demonstrate that incorporating loss derivation into the unsupervised loss of both the discriminator and generator significantly enhances the performance of spam message detection in our work. This notable improvement in our proposed model is due to the use of the Log-Sum-Exp (LSE) function in Eqs. (8) and (15), which effectively addresses underflow and overflow issues.

### 5. Conclusion

In this paper, we presented a conditional semi-supervised data augmentation, i.e., CSSDA. The primary goal of this paper is to address the lack of availability of labeled data for spam detection. The proposed model introduces the fusion of conditional generation and semi-supervised data augmentation. Our approach in data augmentation exploits unlabeled data not only for improving the training inconsistency but also for enhancing the model's final classifier. In addition, we also perform the derivation of unsupervised loss to minimize vanishing gradient. The extensive experimental results indicate that the proposed CSSDA outperforms counterpart models. Study ablation also shows and strengthens the excellent performance of our proposed model. Thus, our proposed CDDA can address the lack of data availability in the spam detection model by exploiting unlabeled data. There is a potential limitation of this paper. That is, our model relies heavily on the content information provided by text data, not non-content-based features such as sender behaviors. Nevertheless, we assert that spam detection remains a persistent challenge, and the aforementioned limitation does not diminish our paper's contribution. In future work, we will expect to extend our model not only for classification tasks but also for more complex tasks such as question answer and others.